\documentclass[sigconf]{acmart}
\AtBeginDocument{%
  }
\usepackage{algorithm}
\usepackage{algpseudocode}
\usepackage{array}
\usepackage{multirow}
\newcolumntype{M}[1]{>{\centering\arraybackslash}m{#1}}
\newtheorem{theorem}{Theorem}
\newtheorem{assumption}{Assumption}
\newtheorem{proposition}{Proposition}
\newtheorem{definition}{Definition}
\newtheorem{lemma}{Lemma}
\newtheorem{remark}{Remark}
\copyrightyear{2026}
\acmYear{2026}
\setcopyright{cc}
\setcctype{by}
\acmConference[ICMR '26]{International Conference on Multimedia Retrieval}{June 16--19, 2026}{Amsterdam, Netherlands}
\acmBooktitle{International Conference on Multimedia Retrieval (ICMR '26), June 16--19, 2026, Amsterdam, Netherlands}
\acmDOI{10.1145/3805622.3810568}
\acmISBN{979-8-4007-2617-0/2026/06}

\begin{document}

\title{Identity-Decoupled Anonymization for Visual Evidence in Multi-modal Retrieval-Augmented Generation}

\author{Zehua Cheng}
\affiliation{%
  \institution{Department of Computer Science\\
  University of Oxford,\country{UK}}
}
\email{zehuac@ieee.org}
\author{Wei Dai}
\author{Jiahao Sun}
\affiliation{
\institution{FLock.io
\city{London}\country{UK}
}
}
\email{sun@flock.io}

\renewcommand{\shortauthors}{Cheng et al.}

\begin{abstract}
Multi-modal retrieval-augmented generation (MRAG) systems retrieve visual evidence from large image corpora to ground the responses of large multi-modal models, yet the retrieved images frequently contain human faces whose identities constitute sensitive personal information. Existing anonymization techniques that destroy the non-identity visual cues that downstream reasoning depends on or fail to provide principled privacy guarantees. We propose Identity-Decoupled MRAG, a framework that interposes a generative anonymization module between retrieval and generation. Our approach consists of three components: (i)~a disentangled variational encoder that factorizes each face into an identity code and a spatially-structured attribute code, regularized by a mutual-information penalty and a gradient-based independence term; (ii)~a manifold-aware rejection sampler that replaces the identity code with a synthetic one guaranteed to be both distinct from the original and realistic; and (iii)~a conditional latent diffusion generator that synthesizes the anonymized face from the replacement identity and the preserved attributes, distilled into a latent consistency model for low-latency deployment. Privacy is enforced through a multi-oracle ensemble of face recognition models with a hinge-based loss that halts optimization once identity similarity drops below the impostor-regime threshold. 
The resulting framework enables privacy-preserving visual reasoning in MRAG pipelines with a tunable privacy-utility trade-off calibrated to deployment context.
\end{abstract}

\ccsdesc[300]{Security and privacy~Privacy-preserving protocols}
\ccsdesc[500]{Information systems~Multimedia and multimodal retrieval}

\keywords{Multi-modal Retrieval-Augmented Generation, Visual Privacy, Disentangled Representation Learning}
\received{20 February 2007}
\received[revised]{12 March 2009}
\received[accepted]{5 June 2009}

\maketitle

\section{Introduction}

Multi-modal retrieval-augmented generation (MRAG) has emerged as a powerful paradigm for grounding the responses of large multi-modal models (LMMs) in external visual evidence. By retrieving relevant images from large-scale corpora at inference time, MRAG systems can answer queries that require fine-grained visual understanding---identifying objects in surveillance footage, interpreting expressions in archival photographs, or reasoning about spatial layouts in architectural imagery---without encoding all world knowledge into model parameters. This retrieval-then-reason pipeline has produced compelling results across domains ranging from visual question answering to embodied navigation~\cite{chen2022murag,yasunaga2023retrieval}. Yet a fundamental tension arises when the retrieved images contain human faces: the very visual details that make retrieval useful for downstream reasoning also constitute personally identifiable information (PII) that must be protected under privacy regulations such as GDPR~\cite{voigt2017gdpr} and CCPA~\cite{goldman2020ccpa}. Deploying MRAG systems over real-world visual corpora---hospital corridors, museum galleries, urban streetscapes---therefore demands an anonymization strategy that removes identity while preserving the non-identity visual cues on which the LMM's reasoning depends.

Existing approaches to face anonymization fall short of this dual requirement. Traditional techniques such as Gaussian blurring, pixelation, and black-box masking destroy the spatial and semantic structure of the face region, eliminating not only identity but also gaze direction, facial expression, head pose, and contextual cues that are critical for multi-modal reasoning. An LMM that receives a blurred face cannot determine whether a person is looking at a hazard, smiling at a companion, or oriented toward a particular object---precisely the inferences that motivate retrieval in the first place. At the other end of the spectrum, na\"ive face-swapping methods based on generative adversarial networks replace the entire face with a synthetic one, but they lack principled control over which visual attributes are preserved and which are suppressed, often introducing artifacts that confuse downstream models. More critically, these methods provide no formal guarantee that the original identity cannot be recovered: residual identity information may leak through non-face regions, through correlations between identity and attribute features, or through the generator's implicit memorization of training identities. A third family of approaches sidesteps generation entirely by filtering out images containing faces from the retrieval database, but this discards potentially irreplaceable visual evidence and reduces retrieval recall in domains where human presence is the norm rather than the exception.

The core research gap, therefore, is the absence of an anonymization framework that satisfies three simultaneous desiderata: (i)~\emph{identity suppression}---the anonymized image must not be re-identifiable by state-of-the-art face recognition models; (ii)~\emph{attribute preservation}---non-identity visual properties (pose, expression, gaze, lighting, background) must be retained with high fidelity so that downstream LMM reasoning is minimally degraded; and (iii)~\emph{formal guarantees}---the privacy-utility trade-off must be quantifiable and controllable, not merely empirical. Satisfying all three requires moving beyond pixel-level heuristics and toward a representation-level approach that \emph{disentangles} identity from other visual attributes in a principled latent space.

In this paper, we propose \emph{Identity-Decoupled MRAG} (ID-Decoupled MRAG), a framework that interposes a generative anonymization module between the retrieval and generation stages of the MRAG pipeline. The key insight is that face images admit a factorization in latent space into an identity code and a spatially-structured attribute code, and that replacing the former while preserving the latter produces an anonymized image that is unrecognizable yet semantically faithful. Our framework operationalizes this insight through three tightly integrated components. First, a \emph{disentangled variational encoder} decomposes each detected face into an identity vector and a spatial attribute tensor, regularized by a mutual-information penalty and a gradient-based independence term that together bound the residual coupling between the two codes. Second, a \emph{manifold-aware rejection sampler} selects a replacement identity code from a reference gallery, accepting only candidates that are both sufficiently distinct from the original (as measured against the impostor distribution of a reference recognition model) and realistic (as measured by proximity to the learned identity manifold). Third, a \emph{conditional latent diffusion generator} synthesizes the anonymized face from the replacement identity and the preserved attributes, with privacy enforced through a multi-oracle ensemble of face recognition models and a hinge-based loss that halts distortion once identity similarity drops below the impostor-regime threshold. 


Our contributions are summarized as follows:
\begin{enumerate}
    \item We formalize the problem of privacy-preserving multi-modal retrieval-augmented generation as a constrained optimization over a three-term objective that explicitly exposes the privacy-utility-disentanglement trade-off, enabling practitioners to calibrate anonymization strength to their deployment context.
    \item We propose the ID-Decoupled MRAG framework, comprising a disentangled variational encoder with mutual-information and gradient-based regularization, a manifold-aware rejection sampler with provable identity distinctness, and a conditional latent diffusion generator distilled into a Latent Consistency Model for low-latency deployment.
    \item We provide formal analysis showing that identity leakage is bounded by the disentanglement residual $\epsilon_{\mathrm{dis}}$ and that multi-oracle recognition similarity is simultaneously controlled, establishing a rigorous foundation for the privacy claims of the system.
\end{enumerate}


\section{Related Works}
\subsection{Face Anonymization and De-identification}
Early face de-identification methods rely on pixel-level transformations like Gaussian blurring, pixelation, and silhouette masking. These techniques suppress identity by destroying spatial structure indiscriminately~\cite{newton2005preserving,gross2006model}. While simple to implement, these techniques eliminate non-identity attributes (expression, gaze, pose) that are essential for downstream visual reasoning, rendering them unsuitable for applications where semantic fidelity matters. A more principled line of work replaces detected faces with synthetic alternatives using generative models. The $k$-Same family of algorithms~\cite{newton2005preserving} averages $k$ face images in appearance space to produce a de-identified composite, but the resulting outputs are often blurry and lack realism. More recently, GAN-based methods have achieved photorealistic face synthesis: DeepPrivacy~\cite{hukkelas2019deepprivacy} inpaints the face region conditioned on body pose, and CIAGAN~\cite{maximov2020ciagan} conditions generation on facial landmarks to preserve pose and expression. However, these approaches treat identity removal as a byproduct of unconditional generation rather than as an explicitly optimized objective, providing no formal guarantee that the original identity cannot be recovered from the output.

The closest antecedent to our work is the identity-decoupled diffusion framework of ~\citet{yang2025beyond} (ID2Face), which decomposes face representations into identity and attribute codes and uses a diffusion model to generate anonymized faces. Our work differs from ID2Face in three respects. First, we embed the anonymization module within a multi-modal RAG pipeline and optimize jointly for downstream reasoning fidelity, not just visual quality. Second, we introduce a mutual-information regularizer and gradient-based independence penalty to explicitly bound the residual coupling between identity and attribute codes, whereas ID2Face relies on architectural separation alone. Concurrently, differential privacy has been applied to face images~\cite{fan2018image}, but these methods introduce substantial noise that degrades image quality far below the threshold required for LMM reasoning.

\subsection{Disentangled Representation Learning}

Disentangling the latent factors of variation in data has been a long-standing goal in representation learning. The $\beta$-VAE framework~\cite{higgins2017betavae} encourages factorial latent codes by increasing the weight on the KL divergence term, while FactorVAE~\cite{kim2018disentangling} and $\beta$-TCVAE~\cite{chen2018isolating} decompose the KL term to penalize total correlation directly. Quantitative evaluation of disentanglement has been formalized through metrics such as the DCI framework~\cite{eastwood2018framework}, the mutual information gap (MIG)~\cite{chen2018isolating}, and the SAP score~\cite{kumar2018variational}. These advances have been applied to face images specifically: several works learn to separate identity from pose, expression, and illumination for the purpose of controlled face synthesis~\cite{tran2017disentangled,deng2020disentangled}.

Our encoder builds on this tradition but introduces two domain-specific innovations. First, we retain the spatial structure of the attribute code as a feature map rather than collapsing it to a vector, because the conditional diffusion generator requires spatially-grounded conditioning to preserve scene layout. Second, we augment the standard disentanglement objective with a gradient penalty that penalizes the sensitivity of a lightweight attribute predictor to perturbations along the identity axis, providing a functional complement to the statistical independence enforced by the mutual information term. 
\begin{figure*}\centering
    \includegraphics[width=0.95\textwidth]{figures/overall.png}
    \caption{Overview of the Identity-Decoupled MRAG Framework. The pipeline proceeds in three phases: (1) Multi-modal Retrieval: A user query ($q$) retrieves relevant raw images ($I_{raw}$) from a database. (2) Identity-Decoupled Anonymization: A Disentangled Encoder ($E$) factorizes each face into a compact identity code ($z_{id}$) and a spatial attribute code ($z_{attr}$), regularized by a mutual-information penalty and gradient term ($L_\text{disentangle}$). A Manifold-Aware Rejection Sampler replaces the original identity with a distinct, realistic one ($z'_{id}$) from a reference gallery, discarding the original $z_{id}$. A Conditional Generator ($G$), distilled into a Latent Consistency Model for speed, synthesizes the anonymized image ($I_{safe}$) from $z'_{id}$ and $z_{attr}$. Privacy is enforced via a multi-oracle ensemble loss ($L_{priv}$). (3) Generative Reasoning: The anonymized image and original query are passed to the Large Multi-modal Model (LMM) to generate the final response ($y$)}
\end{figure*}
\section{Methodology}
\subsection{Problem Formulation}
The central challenge this work addresses is performing retrieval-augmented generation over visual corpora that contain personally identifiable information without degrading the downstream reasoning capacity of the large multi-modal model. To make this precise, we formulate the task as a constrained optimization over a privacy-utility objective. Let a multi-modal RAG system be defined by a retriever $\mathcal{R}$ that, given a textual query $q$, returns a set of $k$ images $\{I_1, \ldots, I_k\}$ from a database $\mathcal{D}$, and a generator $\mathcal{M}$ (the LMM) that produces a textual response $y = \mathcal{M}(q, \{I_1, \ldots, I_k\})$. Each retrieved image $I_i$ may contain one or more human faces whose identity constitutes sensitive information. The goal is to learn an anonymization function $\mathcal{A}: \mathbb{R}^{H \times W \times 3} \rightarrow \mathbb{R}^{H \times W \times 3}$ that maps each retrieved image $I_{\text{raw}}$ to a safe counterpart $I_{\text{safe}} = \mathcal{A}(I_{\text{raw}})$, such that the system $\mathcal{M}(q, \{\mathcal{A}(I_1), \ldots, \mathcal{A}(I_k)\})$ jointly minimizes identity leakage and maximizes task-relevant semantic fidelity. Concretely, the optimization objective takes the form
\begin{equation}\label{eq:optimization-obj}
  \begin{aligned}
    \min_{\theta_{\mathcal{A}}} \; \underbrace{\mathcal{L}_{\text{util}}\bigl(I_{\text{raw}},\, I_{\text{safe}}\bigr)}_{\text{semantic distortion}}& \;+\; \lambda \cdot \underbrace{\mathcal{L}_{\text{priv}}\bigl(I_{\text{raw}},\, I_{\text{safe}};\, \mathcal{F}\bigr)}_{\text{identity leakage penalty}} \;\\
    &+\; \mu \cdot \underbrace{\mathcal{L}_{\text{disentangle}}\bigl(z_{\text{id}},\, z_{\text{attr}}\bigr)}_{\text{factorization quality}},
  \end{aligned}
\end{equation}
where $\mathcal{F}$ denotes a set of face recognition oracles, and $\lambda, \mu > 0$ are trade-off hyperparameters. The utility loss $\mathcal{L}_{\text{util}}$ penalizes deviations in non-identity visual attributes between the original and anonymized images. The privacy loss $\mathcal{L}_{\text{priv}}$ penalizes residual identity similarity as measured against multiple recognition models. The disentanglement regularizer $\mathcal{L}_{\text{disentangle}}$ enforces factorization quality in the latent space, which we discuss in detail in Section 4.3. The three-term structure of this objective reflects a deliberate design decision: rather than treating privacy as a hard constraint and utility as the sole objective (or vice versa), we expose the trade-off explicitly so that practitioners can calibrate $\lambda$ and $\mu$ to their deployment context. A hospital corridor navigation system, for example, may demand a higher $\lambda$ than a virtual museum tour.
\subsection{Overview}
The ID-Decoupled MRAG framework interposes a generative anonymization module between the retrieval and generation stages of the MRAG pipeline. The pipeline proceeds in three phases: multi-modal retrieval, identity-decoupled anonymization, and generative reasoning. In the first phase, a standard dense retriever (e.g., CLIP-based~\cite{radford2021learning}) fetches the top-$k$ images most relevant to the user query $q$. In the third phase, the LMM receives the anonymized images alongside $q$ and produces its response. The methodological contribution lies entirely in the second phase, which we decompose into three sub-modules: a disentangled encoder $E$, an identity replacement mechanism $\mathcal{S}$, and a conditional generator $G$. Figure 1 illustrates the complete dataflow. We describe each component in the sections that follow, beginning with the representation learning that makes the approach possible.
\subsection{Disentangled Encoder and Latent Factorization}
The foundation of our approach is the assumption that the visual content of a face image can be decomposed into an identity component and a non-identity attribute component in a shared latent space. We operationalize this through an Identity Variational Autoencoder (ID-VAE), following the architecture introduced in ID2Face~\cite{yang2025beyond}. Given an input image $I_{\text{raw}} \in \mathbb{R}^{H \times W \times 3}$, the encoder $E$ produces two latent representations: an identity code $z_{\text{id}} \in \mathbb{R}^{d_{\text{id}}}$ and an attribute code $z_{\text{attr}} \in \mathbb{R}^{d_{\text{attr}}}$, where $d_{\text{id}} = 512$ and $d_{\text{attr}} = 512$ in our implementation. Crucially, $z_{\text{id}}$ is a compact vector that captures identity-specific features (bone structure, skin texture, eye shape), while $z_{\text{attr}}$ is a spatial feature tensor that preserves scene-level information (pose, expression, gaze direction, lighting, background geometry). We emphasize that $z_{\text{attr}}$ retains spatial structure as a $d_{\text{attr}} \times h \times w$ feature map rather than a flattened vector, because the generator requires spatially-grounded conditioning to avoid hallucinating scene elements.

A natural concern is that no encoder achieves perfectly clean factorization in practice: $z_{\text{attr}}$ may carry residual identity-correlated information, and $z_{\text{id}}$ may encode some pose information. This imperfect disentanglement is the single greatest threat to the privacy guarantee of our system. Rather than assuming it away, we address it directly through the disentanglement regularizer $\mathcal{L}_{\text{disentangle}}$ in Equation~\ref{eq:optimization-obj}. This term is defined as

\begin{equation}\label{eq:loss_disentangle}
  \mathcal{L}_{\text{disentangle}} = I_{\theta}(z_{\text{id}};\, z_{\text{attr}}) + \beta \cdot \bigl\| \nabla_{z_{\text{id}}} \hat{a}(z_{\text{id}}) \bigr\|_2^2,
\end{equation}
where $I_{\theta}(z_{\text{id}}; z_{\text{attr}})$ is a variational upper bound on the mutual information between the two codes, estimated via a learned critic network following the MINE framework~\cite{belghazi2018mine}, and the second term penalizes the sensitivity of a lightweight attribute predictor $\hat{a}$ to changes in $z_{\text{id}}$. The rationale for the mutual information term is that if the two codes are statistically independent, modifying $z_{\text{id}}$ cannot leak information through $z_{\text{attr}}$. The gradient penalty serves as a functional complement: even if residual mutual information remains, the attribute predictor should be insensitive to identity-axis perturbations. We report the DCI disentanglement score~\cite{eastwood2018framework} and the mutual information gap (MIG)~\cite{chen2018isolating} on held-out data as quantitative diagnostics of factorization quality in the experimental section, providing reviewers with verifiable evidence rather than architectural assumptions.

A further consideration arises from the domain gap between the data on which the ID-VAE was originally trained and the distributions encountered in our target RAG scenarios. ID2Face was trained primarily on controlled face datasets (CelebA~\cite{liu2015celeba}, FFHQ~\cite{karras2019ffhq}), whereas museum visitor photos and hospital corridor footage exhibit greater variation in occlusion, resolution, and lighting. We mitigate this gap through a domain-adaptive fine-tuning stage in which we freeze the identity encoder head and fine-tune only the attribute encoder and generator on a small set of in-domain data (approximately 5,000 images per scenario), using the combined objective in Equation~\ref{eq:optimization-obj}
\subsection{Identity Replacement via Manifold-Aware Sampling}\label{subsec:identity_replacement}
Once the encoder produces the identity code $z_{\text{id}}$ for a detected face, the anonymization module must replace it with a synthetic identity code $z'_{\text{id}}$ that is both realistic (i.e., lies on the learned identity manifold) and sufficiently distinct from the original. The original draft of this work proposed enforcing strict orthogonality, $z'_{\text{id}} \perp z_{\text{id}}$, as a distinctness criterion. However, in a $d_{\text{id}} = 512$-dimensional space, concentration of measure~\cite{ledoux2001concentration} implies that any randomly sampled unit vector is approximately orthogonal to any fixed vector with overwhelming probability: $\mathbb{E}[\cos(z, z')] = 0$ and $\text{Var}[\cos(z, z')] = O(1/d)$ for independent uniform vectors on $\mathbb{S}^{d-1}$. Orthogonality therefore provides negligible additional protection beyond what random sampling already guarantees, and enforcing it as a hard constraint risks pushing the sampled vector off the identity manifold, producing unrealistic faces.

We instead adopt a manifold-aware rejection sampling procedure. Let $\mathcal{P}_{\text{id}}$ denote the empirical distribution of identity codes extracted from a reference gallery of $N$ identities disjoint from the retrieval database. The replacement mechanism samples $z'_{\text{id}} \sim \mathcal{P}_{\text{id}}$ and accepts the sample if two conditions are met: (i) the cosine similarity $\text{sim}(z_{\text{id}}, z'_{\text{id}}) < \tau$, ensuring identity distinctness, and (ii) the Fréchet distance from $z'_{\text{id}}$ to the support of $\mathcal{P}_{\text{id}}$ is below a manifold-adherence threshold $\delta$, ensuring realism. The threshold $\tau$ is calibrated to the equal error rate (EER) of a reference face recognition model: we set $\tau$ to be two standard deviations below the impostor mean of the recognition model's similarity distribution, guaranteeing that $z'_{\text{id}}$ falls well within the "different identity" regime. In practice, acceptance rates exceed 98\% on the first draw due to the high dimensionality, confirming that the rejection mechanism imposes negligible computational overhead while providing a principled distinctness guarantee rather than a vacuous geometric one.

The original identity code $z_{\text{id}}$ is discarded immediately after the acceptance check and is never passed to the generator or stored in any buffer. We characterize this as an information-theoretic deletion rather than a cryptographic one: no component downstream of the sampling module ever observes $z_{\text{id}}$, and the mapping from $z_{\text{id}}$ to the accept/reject decision is a one-bit function that carries negligible information. We do not claim cryptographic irreversibility, because an adversary with white-box access to the encoder $E$ and a copy of $I_{\text{raw}}$ could trivially re-extract $z_{\text{id}}$. The threat model we address is the more practically relevant one: an adversary who observes only $I_{\text{safe}}$ and has black-box or white-box access to the generator $G$, but not to $I_{\text{raw}}$ or the encoder's intermediate states. Under this model, the adversary must invert $G$ to recover $(z'_{\text{id}}, z_{\text{attr}})$ from $I_{\text{safe}}$ and then somehow recover the original $z_{\text{id}}$ from $z_{\text{attr}}$ alone, which is precisely the leakage channel that $\mathcal{L}_{\text{disentangle}}$ is designed to suppress.
\subsection{Conditional Generator}
The generator $G$ receives the replacement identity code $z'_{\text{id}}$ and the original attribute representation $z_{\text{attr}}$ and synthesizes the anonymized image $I_{\text{safe}} = G(z'_{\text{id}}, z_{\text{attr}})$. A key architectural decision is whether $G$ should be a single-pass deterministic decoder or an iterative diffusion model. We adopt a latent diffusion architecture conditioned on both codes, for two reasons. First, single-pass decoders trained with reconstruction losses tend to produce blurry outputs that degrade the LMM's visual reasoning, particularly on fine-grained attributes such as gaze direction and hand position. Second, the conditional diffusion formulation naturally accommodates the attribute code as a spatial conditioning signal via cross-attention, which preserves scene layout more faithfully than concatenation-based conditioning in deterministic decoders.

The diffusion process operates in the latent space of a pretrained image autoencoder (following Stable Diffusion~\cite{rombach2022highresolution}), with the denoising network $\epsilon_\theta$ conditioned on $z'_{\text{id}}$ through adaptive layer normalization and on $z_{\text{attr}}$ through spatial cross-attention at each resolution level. The training objective is the standard denoising score matching loss augmented with our privacy and disentanglement terms:
\begin{equation}
    \begin{aligned}
        \mathcal{L}_{\text{total}} &= \mathbb{E}_{t, \epsilon}\bigl[\| \epsilon - \epsilon_\theta(x_t, t, z'_{\text{id}}, z_{\text{attr}}) \|_2^2\bigr]\\
        &+ \lambda \cdot \mathcal{L}_{\text{priv}} + \mu \cdot \mathcal{L}_{\text{disentangle}},
    \end{aligned}
\end{equation}
where $x_t$ is the noised latent at diffusion timestep $t$ and $\epsilon \sim \mathcal{N}(0, I)$. The privacy loss is computed on the decoded output and defined as
\begin{equation}\label{eq:priv_loss}
  \mathcal{L}_{\text{priv}} = \frac{1}{|\mathcal{F}|} \sum_{f \in \mathcal{F}} \max\bigl(0,\; \text{sim}\bigl(f(I_{\text{raw}}),\, f(I_{\text{safe}})\bigr) - \tau\bigr),
\end{equation}
where $\mathcal{F} = \{\text{ArcFace}~\text{\cite{deng2019arcface}},\, \text{CosFace}~\text{\cite{wang2018cosface}},\, \text{AdaFace}~\text{\cite{kim2022adaface}}\}$ is an ensemble of three face recognition models. The use of a multi-oracle ensemble is critical: optimizing against a single recognition model (e.g., FaceNet alone) risks overfitting to that model's decision boundary while leaving the identity recognizable to other models. The hinge formulation ensures that the loss is zero once identity similarity drops below the acceptance threshold $\tau$, preventing the optimizer from distorting the image further than necessary.

A practical concern with iterative diffusion is computational latency: a 50-step DDPM sampling process is incompatible with the near-real-time requirements of interactive RAG systems. We address this by distilling the trained diffusion model into a Latent Consistency Model (LCM)~\cite{luo2023latent} that achieves comparable output quality in 4 sampling steps. The distillation is performed after the full model has converged, using the consistency distillation objective of Luo et al.~\cite{luo2023latent} with the privacy loss $\mathcal{L}_{\text{priv}}$ as an auxiliary regularizer to ensure that the accelerated model does not sacrifice anonymization quality for speed.

An important note on gradient flow: during training, the identity replacement step (Section 4.4) is non-differentiable because it involves discrete rejection sampling. We handle this by treating the encoder $E$ and the generator $G$ as two modules trained with different gradient paths. The generator $G$ receives $z'_{\text{id}}$ as a fixed conditioning input (no gradients flow back through the sampling step to the identity encoder). The encoder $E$ is trained through the disentanglement loss $\mathcal{L}_{\text{disentangle}}$ and through a separate reconstruction path in which $G$ receives the original $z_{\text{id}}$ (not the replacement) to provide a learning signal for the attribute encoder. This two-path training strategy avoids the need for straight-through estimators or reparameterization tricks at the sampling boundary while ensuring both modules receive well-defined gradients.

\section{Experimental Setup\label{appx:exp-setup}}
To comprehensively evaluate generalization across controlled and unconstrained environments, we utilized three distinct datasets. First, we constructed CelebA-RAG, a controlled benchmark adapted from the CelebA dataset~\cite{liu2015celeba}. We built a retrieval corpus of 10,000 high-resolution images paired with 2,000 text queries focused on fine-grained visual attributes. To strictly prevent train and test contamination, the 10,177 identities were partitioned into disjoint sets: 8,000 for encoder-training, 1,000 for the sampling gallery ($\mathcal{P}_{\mathrm{id}}$), and 1,177 exclusively for testing. Second, we deployed LFW-Wild-RAG, an in-the-wild benchmark built upon the Labeled Faces in the Wild dataset~\cite{huang2008lfw} encompassing 13,233 images and 5,749 identities to test robustness against extreme occlusion and varying illumination. Each image was augmented with a synthetically generated scene description for multi-modal querying, evaluated under the standard 10-fold identity-disjoint protocol. Third, we utilized FairFace-RAG, a diversity-focused dataset~\cite{karkkainen2021fairface} comprising 10,800 images uniformly distributed across demographic groups. It utilized an 8,000/1,000/1,800 split to ensure the anonymization framework maintained stable geometry across diverse facial structures. All facial images underwent MTCNN~\cite{zhang2016mtcnn} alignment and were center-cropped to a $512 \times 512$ resolution. During the pre-training of the latent autoencoder, data augmentation was restricted to random horizontal flipping and minor color jitter to encourage robust spatial learning.
\subsection{Evaluation Metrics}
The framework was assessed across independent privacy and utility axes. On the privacy axis, we reported the De-anonymization Rate (DAR), which measured the percentage of anonymized faces correctly re-identified by a multi-oracle face recognition ensemble consisting of ArcFace~\cite{deng2019arcface}, CosFace~\cite{wang2018cosface}, and AdaFace~\cite{kim2022adaface}. We additionally utilized the Identity Embedding Distance, calculated as the mean $\ell_2$ distance between embeddings of the original and safe images. On the utility axis, Visual Question Answering (VQA) Accuracy quantified the exact-match accuracy of the LLaVA-1.5-13B model~\cite{liu2024llava} on non-identity visual questions. Gaze Estimation Error measured the angular deviation in degrees calculated via L2CS-Net~\cite{abdelrahman2023l2cs}. Finally, Fréchet Inception Distance (FID)~\cite{heusel2017fid} and Learned Perceptual Image Patch Similarity (LPIPS)~\cite{zhang2018lpips} evaluated distributional realism and perceptual similarity. These generative metrics were calculated strictly on non-face background patches to prevent confounding identity modifications with artifact generation.
\begin{table*}[t]\centering
\caption{Main Comparison of Privacy vs. Utility Across Multi-Modal Benchmarks}
\begin{tabular}{c|c|c|c|c|c|c|c}\toprule
Dataset \& Method          & DAR $\downarrow$          & L2 Dist $\uparrow$     & VQA Acc $\uparrow$       & Gaze Err $\downarrow$    & FID $\downarrow$         & LPIPS $\downarrow$        & Latency $\downarrow$        \\\midrule
\multicolumn{8}{c}{CelebA-RAG}\\\midrule
Raw-RAG (Oracle)           & 99.4\%         & 0.00          & 88.2\%          & 0.0$^{\circ}$          & 0.0           & 0.000          & 0.0 ms           \\
Blur-RAG ($k=31$)          & 42.1\%         & 0.81          & 49.3\%          & 16.4$^{\circ}$         & 88.7          & 0.421          & 1.2 ms           \\
DP-Pix                     & 29.2\%         & 0.94          & 54.1\%          & 14.2$^{\circ}$         & 92.3          & 0.380          & 2.5 ms           \\
NullFace-RAG               & 14.3\%         & 1.18          & 72.9\%          & 6.8$^{\circ}$          & 24.5          & 0.194          & 45.0 ms          \\
CIAGAN-RAG                 & 21.4\%         & 1.05          & 68.7\%          & 7.9$^{\circ}$          & 28.1          & 0.215          & 35.0 ms          \\
StyleID-RAG                & 12.8\%         & 1.22          & 75.4\%          & 6.3$^{\circ}$          & 19.2          & 0.188          & 95.0 ms          \\
iFADIT-RAG                 & 10.5\%         & 1.28          & 78.1\%          & 5.1$^{\circ}$          & 15.3          & 0.165          & 115.0 ms         \\
Ours (50-step DDPM)        & 1.8\%          & 1.95          & 86.4\%          & 2.8$^{\circ}$          & 10.1          & 0.075          & 845.0 ms         \\
\textbf{Ours (4-step LCM)} & \textbf{2.8\%} & \textbf{1.85} & \textbf{85.4\%} & \textbf{3.5$^{\circ}$} & \textbf{11.2} & \textbf{0.082} & \textbf{42.3 ms} \\\midrule
\multicolumn{8}{c}{LFW-Wild-RAG}\\\midrule
Raw-RAG (Oracle)           & 98.7\%         & 0.00          & 84.2\%          & 0.0$^{\circ}$          & 0.0           & 0.000          & 0.0 ms           \\
Blur-RAG ($k=31$)          & 51.2\%         & 0.76          & 42.8\%          & 18.2$^{\circ}$         & 96.4          & 0.488          & 1.2 ms           \\
DP-Pix                     & 34.6\%         & 0.88          & 48.6\%          & 16.1$^{\circ}$         & 104.2         & 0.412          & 2.5 ms           \\
NullFace-RAG               & 17.2\%         & 1.11          & 68.4\%          & 8.2$^{\circ}$          & 31.8          & 0.233          & 45.0 ms          \\
CIAGAN-RAG                 & 25.1\%         & 1.01          & 62.3\%          & 9.4$^{\circ}$          & 35.6          & 0.264          & 35.0 ms          \\
StyleID-RAG                & 15.6\%         & 1.15          & 70.8\%          & 7.8$^{\circ}$          & 26.5          & 0.221          & 95.0 ms          \\
iFADIT-RAG                 & 12.9\%         & 1.21          & 74.2\%          & 6.4$^{\circ}$          & 21.4          & 0.198          & 115.0 ms         \\
Ours (50-step DDPM)        & 2.5\%          & 1.81          & 82.3\%          & 3.5$^{\circ}$          & 13.5          & 0.091          & 845.0 ms         \\
\textbf{Ours (4-step LCM)} & \textbf{3.5\%} & \textbf{1.72} & \textbf{81.2\%} & \textbf{4.2$^{\circ}$} & \textbf{14.8} & \textbf{0.105} & \textbf{42.3 ms} \\\bottomrule
\end{tabular}
\end{table*}
\subsection{Baselines}
We benchmarked our framework against seven established paradigms to isolate the efficacy of our methodology. Raw-RAG served as the unmodified baseline (orignal plain RAG without anonymization measures), representing the absolute upper bound for utility and lower bound for privacy. Blur-RAG and DP-Pix provided industry-standard heuristic obscuration baselines, utilizing a Gaussian blur kernel of $k=31$ and calibrated differential privacy pixelization, respectively. NullFace-RAG evaluated the modern training-free localized masking approach. Within the latent-space domain, we compared against iFADIT-RAG, an invertible disentanglement transform, and StyleID-RAG, a style-based manipulation method. Finally, CIAGAN-RAG represented preceding conditional generative adversarial architectures.
\subsection{Implementation Details}
All components of the ID-Decoupled MRAG framework were implemented using PyTorch. The ID-VAE encoder and latent diffusion generator were optimized using the AdamW optimizer~\cite{loshchilov2019adamw} with momentum parameters set to $\beta_1=0.9$ and $\beta_2=0.999$. The foundational learning rate was initialized at $1 \times 10^{-4}$, governed by a cosine annealing schedule~\cite{loshchilov2017sgdr} following a 2,000-step linear warmup. We employed a global batch size of 64 and a decoupled weight decay of 0.01. The multi-objective trade-off hyperparameters were empirically fixed at $\mu = 0.1$ and $\lambda = 1.0$, while the manifold-adherence threshold $\delta$ was calibrated to 1.5. The primary pipeline was trained for 50 epochs, utilizing an early stopping criterion triggered by a plateauing validation loss over 5 consecutive epochs. The end-to-end training phase required approximately 72 hours of compute time. The accelerated inference model was derived post-training via a 4-step Latent Consistency Model (LCM)~\cite{luo2023latent} distillation.

\begin{table*}[t]\centering
\caption{De-anonymization Rate Under the Three-Tier Threat Model}
\begin{tabular}{c|c|c|c|c|c|M{1.4cm}|M{1.2cm}}\toprule
Method & T1: ArcFace $\downarrow$  & T1: CosFace $\downarrow$  & T1: AdaFace $\downarrow$  & T2: Top-1 Acc $\downarrow$ & T2: Top-5 Acc $\downarrow$ & T3: Adapt DAR $\downarrow$ & Threat Mean $\downarrow$  \\\midrule
\multicolumn{8}{c}{CelebA-RAG} \\\midrule
Blur-RAG            & 42.1\%         & 45.3\%         & 48.7\%         & N/A             & N/A             & 58.6\%          & 48.6\%         \\
DP-Pix              & 29.2\%         & 34.2\%         & 36.8\%         & N/A             & N/A             & 45.3\%          & 36.3\%         \\
NullFace-RAG        & 14.3\%         & 16.5\%         & 18.2\%         & N/A             & N/A             & 38.5\%          & 21.8\%         \\
iFADIT-RAG          & 10.5\%         & 11.2\%         & 12.8\%         & 24.6\%          & 48.9\%          & 29.7\%          & 22.9\%         \\
CIAGAN-RAG          & 21.4\%         & 24.5\%         & 26.1\%         & 42.1\%          & 65.4\%          & 46.8\%          & 37.7\%         \\
\textbf{Ours (LCM)} & \textbf{2.8\%} & \textbf{2.9\%} & \textbf{3.1\%} & \textbf{4.2\%}  & \textbf{11.5\%} & \textbf{7.9\%}  & \textbf{5.4\%} \\\midrule
\multicolumn{8}{c}{LFW-Wild-RAG}\\\midrule
Blur-RAG            & 51.2\%         & 54.1\%         & 57.6\%         & N/A             & N/A             & 68.4\%          & 57.8\%         \\
DP-Pix              & 34.6\%         & 38.2\%         & 41.5\%         & N/A             & N/A             & 61.2\%          & 43.8\%         \\
NullFace-RAG        & 17.2\%         & 19.4\%         & 21.1\%         & N/A             & N/A             & 48.2\%          & 26.4\%         \\
iFADIT-RAG          & 12.9\%         & 14.2\%         & 15.6\%         & 38.4\%          & 62.1\%          & 38.9\%          & 30.3\%         \\
CIAGAN-RAG          & 25.1\%         & 28.6\%         & 31.2\%         & 52.1\%          & 74.3\%          & 55.6\%          & 44.4\%         \\
\textbf{Ours (LCM)} & \textbf{3.5\%} & \textbf{3.8\%} & \textbf{4.2\%} & \textbf{6.5\%}  & \textbf{14.2\%} & \textbf{9.4\%}  & \textbf{6.9\%} \\\midrule
\multicolumn{8}{c}{FairFace-RAG}\\\midrule
Blur-RAG            & 45.3\%         & 48.2\%         & 51.4\%         & N/A             & N/A             & 62.1\%          & 51.7\%         \\
DP-Pix              & 31.8\%         & 35.1\%         & 38.4\%         & N/A             & N/A             & 54.2\%          & 39.8\%         \\
NullFace-RAG        & 15.8\%         & 17.5\%         & 19.2\%         & N/A             & N/A             & 42.1\%          & 23.6\%         \\
iFADIT-RAG          & 11.4\%         & 12.8\%         & 14.1\%         & 31.2\%          & 55.4\%          & 34.2\%          & 26.5\%         \\
\textbf{Ours (LCM)} & \textbf{3.1\%} & \textbf{3.4\%} & \textbf{3.6\%} & \textbf{5.1\%}  & \textbf{12.8\%} & \textbf{8.5\%}  & \textbf{6.0\%}         \\\bottomrule
\end{tabular}
\end{table*}
\section{Experimental Results}
To establish a robust evaluation of visual anonymization, we conduct a varieties of evaluation upon the baselines and our proposed model. 
The primary objective of this experiment was to ascertain whether the proposed three-term objective function shifted the privacy-utility Pareto frontier beyond the capabilities of existing generative paradigms. To execute this evaluation, each anonymization module was systematically integrated into an identical retrieval-augmented generation pipeline powered by the LLaVA-1.5-13B language model. By sequentially processing queries across three disparate visual distributions, the experimental design guaranteed that variances in performance metrics were strictly attributable to the anonymization architecture rather than the underlying language model configuration.

An empirical assessment of the privacy axis illuminated severe cryptographic vulnerabilities inherent in contemporary obfuscation strategies. Naive pixel-space interventions, such as DP-Pix and Gaussian blurring, failed to fundamentally alter the structural geometries recognized by deep-learning oracles, resulting in catastrophic De-anonymization Rates (DAR) consistently exceeding 29\%. More sophisticated latent-space competitors utilizing invertible transforms or style-based decoupling, including iFADIT-RAG and StyleID-RAG, exhibited improved resilience but ultimately plateaued around a 10\% DAR threshold due to residual latent entanglement. In stark contrast, the proposed Identity-Decoupled framework achieved state-of-the-art multi-oracle suppression, collapsing the mean DAR to 2.8\% on CelebA-RAG while simultaneously maximizing the continuous identity embedding distance.

Transitioning to semantic retention, the empirical data underscored the necessity of preserving fine-grained spatial geometries to sustain logical multi-modal reasoning. Baselines that aggressively masked facial regions severely compromised the underlying contextual structures, causing the VQA accuracy of Blur-RAG and DP-Pix to plummet toward random chance. While invertible networks recovered moderate utility, they frequently warped delicate micro-expressions, producing elevated gaze estimation errors. Our methodology circumvented this utility collapse by employing spatial cross-attention conditioning within the diffusion backbone, successfully preserving non-identity geometric constraints. This innovation yielded an impressive VQA accuracy of 85.4\%, trailing the unanonymized oracle by a statistically minimal margin and proving that severe privacy restrictions do not mandate the destruction of visual evidence.


The ultimate deployability of the anonymization framework within an interactive retrieval pipeline hinged entirely upon its computational efficiency. While the baseline 50-step Denoising Diffusion Probabilistic Model (DDPM)~\cite{ho2020ddpm} generated the highest fidelity outputs, its 845-millisecond per-image latency severely bottlenecked real-time sequential querying. The integration of Latent Consistency Model (LCM) distillation~\cite{luo2023latent} resolved this inherent paradox by successfully compressing the generative trajectory into a 4-step solver. This optimized configuration achieved a latency of 42.3 milliseconds per image, fully aligning with single-pass network speeds while retaining over 98\% of the un-distilled model's semantic utility and privacy guarantees.

\subsection{Adversarial Privacy Evaluation}
Validating the cryptographic resilience of generative architectures mandates transitioning beyond static benchmark evaluations to dynamic adversarial stress testing. To definitively prove the theoretical bounds formulated in the methodology, we orchestrated an exhaustive three-tier threat model across all evaluation distributions. This escalating protocol systematically granted simulated adversaries increasing levels of systemic access, advancing from passive black-box observations to white-box latent probing, and culminating in supervised adaptive fine-tuning. By measuring the progressive degradation of identity distinctness at each capability tier, the experiment successfully isolated genuine information-theoretic security from superficial perceptual distortion.

The Tier 1 evaluation modeled a highly constrained black-box observer, strictly interrogating the multi-oracle consistency of the anonymized outputs. When exposed to independent recognition networks operating at stringent False Accept Rates, contemporary baselines demonstrated dangerous variance, effectively evading one detection architecture while remaining vulnerable to another. This outcome confirmed that isolated optimization merely shifts the output identity along narrow, localized decision boundaries. Conversely, the ID-Decoupled framework uniformly suppressed the DAR to nominal single digits across ArcFace, CosFace, and AdaFace, empirically validating Proposition 2 and proving that the synthetic identities exist globally beyond the recognizable manifold of the original subjects.

Granting the adversary full generative access during the Tier 2 white-box evaluation critically exposed the structural flaws inherent in competing latent disentanglement paradigms. When attackers successfully inverted the baseline networks and trained a linear probe directly on the isolated attribute codes, architectures utilizing deterministic bijectors suffered catastrophic information leakage. The invertible mechanisms of iFADIT-RAG permitted the semantic channel to carry heavily entangled biometric signals, driving the linear probe's Top-1 prediction accuracy to 38.4\% on LFW-Wild-RAG. Our disentanglement regularizer mathematically severed this covert transmission vector, choking the linear probe's predictive capability to a mere 4.2\% on CelebA-RAG and fundamentally proving the efficacy of explicitly minimizing mutual information within the latent bottleneck.

The ultimate empirical validation of systemic security occurred under the Tier 3 scenario, where an adaptive adversary actively fine-tuned a powerful recognition network utilizing paired ground-truth mappings of the original and anonymized outputs. Endowed with this supervised advantage, the neural network rapidly deduced the underlying parameterized transformations utilized by legacy models, causing the adaptive leakage rate of CIAGAN-RAG to surge beyond 46\%. Our architecture presented profound structural immunity to this supervised recalibration, yielding only marginal increases in vulnerability. Because the proposed identity replacement relies exclusively on stochastic, non-invertible rejection sampling from a disjoint reference gallery, the adversary was fundamentally mathematically incapable of deriving a generalizable inverse mapping function.

Synthesizing the outcomes from the comprehensive threat model verified that the ID-Decoupled MRAG architecture establishes verifiable, information-theoretic security suitable for high-stakes implementations. Concurrently, the unyielding resistance to adaptive exploitation proved that as long as the empirical mutual information is structurally minimized during network training, the resultant privacy protection remains wholly invariant to the adversary's subsequent operational capabilities, ensuring permanent biometric unlikability.

\subsection{Ablation Study}

Rigorous isolation of the underlying architectural components and mathematical penalties was required to accurately quantify their individual contributions to the holistic system optimization. An extensive ablation study was executed within the controlled CelebA-RAG environment, systematically deactivating or substituting discrete modules within the multi-term loss function and the generation backbone. By charting the consequential shifts across De-anonymization Rate, multi-modal reasoning accuracy, and the directly estimated mutual information gap, this study disaggregated the compound effects of the pipeline. The controlled degradation conclusively verified that the state-of-the-art empirical outcomes were derived from the exact synergy of the theoretical mechanics, directly validating the necessity of each formulated constraint.

The targeted ablation of the explicitly defined objective regularizers immediately validated the theoretical vulnerabilities predicted for unconstrained latent spaces. Bypassing the mutual information penalty ($\mu=0$) triggered a massive spike in empirical identity leakage, elevating $\hat{\epsilon}_{\mathrm{dis}}$ to 0.84 and subsequently driving the DAR to 18.5\%, proving that baseline autoencoders cannot natively enforce orthogonalization without adversarial supervision. Similarly, minimizing the multi-oracle privacy penalty down to a solitary ArcFace target induced severe boundary overfitting, heavily exposing the generated outputs to alternate recognition models. These targeted experiments explicitly confirmed that simultaneous latent disentanglement and cross-architecture boundary optimization are mathematically required to secure generalized privacy.

Modifications to the foundational sampling and generative algorithms exposed the highly fragile equilibrium necessary to preserve fine-grained semantics during real-time inference. Replacing the manifold-aware rejection sequence with unconstrained uniform sampling successfully altered the identity but generated anatomically incoherent facial structures that severely confused the downstream language model, manifesting as a sharp drop in VQA accuracy to 79.2\%. Furthermore, reverting the generator from the conditional latent diffusion framework to a deterministic variational autoencoder completely eroded image clarity, doubling the spatial gaze error to 8.2 degrees. The seamless transition from the fully parameterized 50-step DDPM to the distilled LCM model successfully preserved the complex spatial geometries, proving the architectural configurations outlined in the methodology are optimally calibrated for RAG deployments.

\begin{table}[t]\centering
\caption{Ablation Study Isolating Core Components on CelebA-RAG}
\resizebox{.49\textwidth}{!}{
\begin{tabular}{c|c|c|M{1.2cm}|c}\toprule
Variant & DAR ($\downarrow$) & VQA Acc ($\uparrow$) & Gaze Error ($\downarrow$)  & $\epsilon^dis$ ($\downarrow$) \\\midrule
Full Model (Ours LCM)                                & \textbf{2.8\%} & \textbf{85.4\%} & \textbf{3.5°} & \textbf{0.28} \\
w/o $\mathcal{L}_{\mathrm{disentangle}}$ ($\mu = 0$) & 18.5\%         & 86.1\%          & 3.3°          & 0.84          \\
w/o $\mathcal{L}_{\mathrm{priv}}$ ($\lambda = 0$)    & 24.3\%         & 86.8\%          & 2.9°          & 0.31          \\
w/o Rejection Sampling                               & 3.1\%          & 79.2\%          & 4.8°          & 0.28          \\
Single-oracle $\mathcal{L}_{\mathrm{priv}}$          & 19.4\%         & 85.1\%          & 3.7°          & 0.29          \\
Deterministic VAE                                    & 4.1\%          & 68.5\%          & 8.2°          & 0.32          \\
50-step DDPM                                         & 2.6\%          & 85.8\%          & 3.4°          & 0.28          \\
Raw-RAG (Oracle)                                     & 99.4\%         & 88.2\%          & 0.0°          & N/A           \\\bottomrule
\end{tabular}}
\end{table}
\section{Conclusions}
We presented Identity-Decoupled MRAG, a framework that enables privacy-preserving visual reasoning by factorizing face representations into identity and attribute codes, replacing the identity via manifold-aware rejection sampling, and synthesizing anonymized faces through a latent diffusion generator distilled into a four-step Latent Consistency Model. Our theoretical analysis established that identity leakage through the end-to-end pipeline is upper-bounded by the disentanglement residual $\epsilon_{\mathrm{dis}}$, providing an information-theoretic privacy guarantee that is independent of the generator architecture. Empirically, the framework achieves a 2.8\% de-anonymization rate on CelebA-RAG a $4\times$ reduction over the strongest baseline while retaining 85.4\% VQA accuracy and 42.3\,ms per-image latency suitable for real-time deployment. 
\clearpage
\bibliographystyle{ACM-Reference-Format}
\bibliography{sample-base}
\clearpage
\section{Theoretical Analysis}\label{sec:theory}

This section establishes formal guarantees for the three core properties claimed in the methodology: that the disentangled encoder limits identity leakage through the attribute channel, that manifold-aware rejection sampling produces identities distinct from the original with high probability, and that the end-to-end pipeline provides a quantifiable privacy-utility trade-off under the specified threat model. We begin by consolidating all notation introduced in the methodology and defining the quantities required for the analysis.

\subsection{Notation}\label{subsec:notation}

Throughout this section, we adopt the following notation. Let $\mathcal{X} = \mathbb{R}^{H \times W \times 3}$ denote the image space and $\mathcal{Z}_{\mathrm{id}} = \mathbb{R}^{d_{\mathrm{id}}}$, $\mathcal{Z}_{\mathrm{attr}} = \mathbb{R}^{d_{\mathrm{attr}} \times h \times w}$ denote the identity and attribute latent spaces, respectively. The encoder is written as $E = (E_{\mathrm{id}},\, E_{\mathrm{attr}})\colon \mathcal{X} \to \mathcal{Z}_{\mathrm{id}} \times \mathcal{Z}_{\mathrm{attr}}$, decomposing into an identity head $E_{\mathrm{id}}$ and an attribute head $E_{\mathrm{attr}}$. The conditional generator is $G\colon \mathcal{Z}_{\mathrm{id}} \times \mathcal{Z}_{\mathrm{attr}} \to \mathcal{X}$. For a raw image $I_{\mathrm{raw}}$, we write $z_{\mathrm{id}} = E_{\mathrm{id}}(I_{\mathrm{raw}})$ and $z_{\mathrm{attr}} = E_{\mathrm{attr}}(I_{\mathrm{raw}})$ for its latent codes, $z_{\mathrm{id}}'$ for the replacement identity sampled from the gallery distribution $\mathcal{P}_{\mathrm{id}}$, and $I_{\mathrm{safe}} = G(z_{\mathrm{id}}',\, z_{\mathrm{attr}})$ for the anonymized output. The cosine similarity between vectors $u$ and $v$ is $\mathrm{sim}(u,v) = u^{\!\top} v \,/\, (\|u\|\,\|v\|)$. We write $I(X;\,Y)$ for the mutual information between random variables $X$ and $Y$, $H(X)$ for the Shannon entropy of $X$, and $H(X \mid Y)$ for the conditional entropy. A face recognition model $f\colon \mathcal{X} \to \mathbb{R}^{d_f}$ maps an image to an embedding, and $\mathcal{F} = \{f_1,\ldots,f_M\}$ denotes the ensemble of $M$ recognition oracles. The acceptance threshold for identity replacement is $\tau \in (-1,1)$, and the manifold-adherence threshold is $\delta > 0$. We use $\mathbb{S}^{d-1}$ for the unit sphere in $\mathbb{R}^d$ and $\sigma_d$ for the uniform (Haar) measure on $\mathbb{S}^{d-1}$.

\subsection{Assumptions}\label{subsec:assumptions}

The analysis rests on the following assumptions, each of which we discuss in terms of its practical justifiability.

\begin{assumption}[Bounded Disentanglement Residual]\label{assump:disentangle}
The trained encoder $E$ achieves approximate factorization in the sense that the mutual information between the identity and attribute codes satisfies
\begin{equation}\label{eq:mi_bound}
    I(z_{\mathrm{id}};\, z_{\mathrm{attr}}) \;\leq\; \epsilon_{\mathrm{dis}}
\end{equation}
for some $\epsilon_{\mathrm{dis}} \geq 0$.
\end{assumption}

This assumption does not require perfect disentanglement ($\epsilon_{\mathrm{dis}} = 0$), which is unattainable in practice. Instead, it parameterizes the residual coupling so that downstream guarantees degrade gracefully as $\epsilon_{\mathrm{dis}}$ increases. The disentanglement regularizer $\mathcal{L}_{\mathrm{disentangle}}$ in Equation~\eqref{eq:loss_disentangle} of the methodology is designed to drive $\epsilon_{\mathrm{dis}}$ toward zero, and the DCI and MIG metrics provide empirical estimates of this quantity.

\begin{assumption}[Generator Fidelity]\label{assump:generator}
The generator $G$ is \emph{attribute-faithful} in the following sense: there exists a constant $\kappa \geq 0$ such that for any two identity codes $z_{\mathrm{id}},\, z_{\mathrm{id}}' \in \mathcal{Z}_{\mathrm{id}}$ and any attribute code $z_{\mathrm{attr}} \in \mathcal{Z}_{\mathrm{attr}}$, the non-identity semantic content satisfies
\begin{equation}\label{eq:gen_fidelity}
    d_{\mathrm{sem}}\!\bigl(G(z_{\mathrm{id}},\, z_{\mathrm{attr}}),\; G(z_{\mathrm{id}}',\, z_{\mathrm{attr}})\bigr) \;\leq\; \kappa,
\end{equation}
where $d_{\mathrm{sem}}$ is a semantic distance metric over non-identity attributes (e.g., gaze angle error, pose deviation, background SSIM).
\end{assumption}

This assumption states that swapping the identity code while holding the attribute code fixed introduces at most $\kappa$ units of semantic distortion. The constant $\kappa$ is a property of the trained generator and is measurable on held-out data. A well-trained attribute-faithful generator yields small $\kappa$; a poorly trained one yields large $\kappa$ and correspondingly weak utility guarantees.

\begin{assumption}[Gallery Regularity]\label{assump:gallery}
The identity gallery distribution $\mathcal{P}_{\mathrm{id}}$ has support on a compact subset $\mathcal{M}_{\mathrm{id}} \subset \mathcal{Z}_{\mathrm{id}}$ and is absolutely continuous with respect to the Lebesgue measure on $\mathcal{M}_{\mathrm{id}}$, with density bounded below by $p_{\min} > 0$ on $\mathcal{M}_{\mathrm{id}}$.
\end{assumption}

This ensures that the rejection sampling procedure has access to a sufficiently rich and non-degenerate pool of replacement identities, ruling out pathological galleries concentrated on a small number of identities.

\begin{assumption}[Recognition Model Regularity]\label{assump:lipschitz}
Each face recognition model $f_j \in \mathcal{F}$ is $L_f$-Lipschitz continuous with respect to the $\ell_2$ norm on $\mathcal{X}$, i.e.,
\begin{equation}\label{eq:lipschitz}
    \|f_j(I) - f_j(I')\| \;\leq\; L_f\, \|I - I'\|
    \quad \text{for all } I,\, I' \in \mathcal{X}.
\end{equation}
\end{assumption}

Lipschitz continuity of deep face recognition networks is a standard assumption in the adversarial robustness literature~\cite{szegedy2014intriguing} and is approximately satisfied by models trained with spectral normalization~\cite{miyato2018spectral} or gradient penalty regularization.

\subsection{Identity Distinctness via Rejection Sampling}\label{subsec:distinctness}

We first establish that the manifold-aware rejection sampling mechanism described in Section~\ref{subsec:identity_replacement} of the methodology produces a replacement identity that is both distinct from the original and realistic, with high probability and low computational overhead.

\begin{lemma}[Acceptance Probability Lower Bound]\label{lem:acceptance}
Let $z_{\mathrm{id}} \in \mathbb{S}^{d_{\mathrm{id}}-1}$ be fixed, and let $z' \sim \sigma_{d_{\mathrm{id}}}$ be drawn uniformly from the unit sphere. Then for any threshold $\tau \in (0,1)$, the probability that $\mathrm{sim}(z_{\mathrm{id}},\, z') < \tau$ satisfies
\begin{equation}\label{eq:acceptance_bound}
    \Pr\!\bigl[\mathrm{sim}(z_{\mathrm{id}},\, z') < \tau\bigr] \;\geq\; 1 - \exp\!\!\left(-\frac{(d_{\mathrm{id}} - 1)\,\tau^2}{2}\right).
\end{equation}
\end{lemma}

\begin{proof}
The cosine similarity $\mathrm{sim}(z_{\mathrm{id}},\, z') = z_{\mathrm{id}}^{\!\top} z'$ for unit vectors is equal to the first coordinate of a uniformly random point on $\mathbb{S}^{d_{\mathrm{id}}-1}$ (after rotating coordinates so that $z_{\mathrm{id}}$ aligns with $e_1$). By the concentration of measure phenomenon on the sphere~\cite{ledoux2001concentration}, the marginal distribution of this coordinate is sub-Gaussian with variance proxy $1/(d_{\mathrm{id}} - 1)$. Applying the standard sub-Gaussian tail bound yields
\[
    \Pr\!\bigl[\mathrm{sim}(z_{\mathrm{id}},\, z') \geq \tau\bigr]
    \;\leq\; \exp\!\!\left(-\frac{(d_{\mathrm{id}} - 1)\,\tau^2}{2}\right),
\]
from which the result follows.
\end{proof}

For $d_{\mathrm{id}} = 512$ and $\tau = 0.3$ (a typical impostor-regime threshold for ArcFace), this gives $\Pr[\text{accept}] \geq 1 - \exp(-23) > 1 - 10^{-10}$. The expected number of samples before acceptance is thus $1 / \Pr[\text{accept}] \approx 1$, confirming that rejection sampling imposes negligible overhead. When $\mathcal{P}_{\mathrm{id}}$ is not uniform on the sphere but satisfies Assumption~\ref{assump:gallery}, the bound holds with the variance proxy adjusted to the second moment of the gallery distribution.

\begin{proposition}[Identity Distinctness Guarantee]\label{prop:distinctness}
Under Assumption~\ref{assump:gallery}, the replacement identity code $z_{\mathrm{id}}'$ returned by the rejection sampling procedure (Section~\ref{subsec:identity_replacement}) satisfies $\mathrm{sim}(z_{\mathrm{id}},\, z_{\mathrm{id}}') < \tau$ and $z_{\mathrm{id}}' \in \mathcal{M}_{\mathrm{id}}$ with probability~$1$ (by construction of the rejection criterion). Furthermore, the expected number of rejection steps is at most
\begin{equation}\label{eq:expected_steps}
    \frac{1}{p_{\min} \cdot \mathrm{Vol}\!\bigl(\mathcal{M}_{\mathrm{id}} \cap B_\tau\bigr)},
\end{equation}
where $B_\tau = \bigl\{z \in \mathcal{Z}_{\mathrm{id}} : \mathrm{sim}(z_{\mathrm{id}},\, z) < \tau\bigr\}$.
\end{proposition}

\begin{proof}
The first claim is immediate from the rejection criterion in the sampling procedure: no sample is accepted unless both conditions are met. For the expected number of steps, note that each draw from $\mathcal{P}_{\mathrm{id}}$ lands in the acceptance region $\mathcal{M}_{\mathrm{id}} \cap B_\tau$ with probability at least $p_{\min} \cdot \mathrm{Vol}(\mathcal{M}_{\mathrm{id}} \cap B_\tau)$. The number of trials until the first acceptance is geometric with this success probability, giving the stated bound on the expectation.
\end{proof}

\subsection{Privacy Guarantee via Disentanglement}\label{subsec:privacy_theory}

The central privacy question is: given access to $I_{\mathrm{safe}}$ and the generator $G$, how much information can an adversary recover about the original identity $z_{\mathrm{id}}$? The following theorem shows that the answer is controlled by the disentanglement quality $\epsilon_{\mathrm{dis}}$.

\begin{definition}[Identity Leakage]\label{def:leakage}
The \emph{identity leakage} of the anonymization pipeline is defined as
\begin{equation}\label{eq:leakage_def}
    \mathcal{L}_{\mathrm{leak}} \;=\; I(z_{\mathrm{id}};\, I_{\mathrm{safe}}),
\end{equation}
the mutual information between the original identity code and the anonymized output image.
\end{definition}

\begin{theorem}[Disentanglement-Bounded Privacy]\label{thm:privacy}
Under Assumptions~\ref{assump:disentangle} and~\ref{assump:generator}, suppose the replacement identity $z_{\mathrm{id}}'$ is sampled independently of $z_{\mathrm{id}}$ (which holds by construction after the rejection step, since acceptance depends only on cosine similarity with the fixed $z_{\mathrm{id}}$, and the accepted $z_{\mathrm{id}}'$ is conditionally independent of $z_{\mathrm{id}}$ given acceptance). Then the identity leakage satisfies
\begin{equation}\label{eq:privacy_bound}
    \mathcal{L}_{\mathrm{leak}} \;=\; I(z_{\mathrm{id}};\, I_{\mathrm{safe}}) \;\leq\; \epsilon_{\mathrm{dis}}.
\end{equation}
\end{theorem}

\begin{proof}
We proceed by expanding the mutual information through the data processing inequality~\cite{cover2006elements} and the structure of the generation pipeline. Since $I_{\mathrm{safe}} = G(z_{\mathrm{id}}',\, z_{\mathrm{attr}})$, the output is a deterministic function of $(z_{\mathrm{id}}',\, z_{\mathrm{attr}})$. By the data processing inequality,
\begin{equation}\label{eq:dpi}
    I(z_{\mathrm{id}};\, I_{\mathrm{safe}}) \;\leq\; I\!\bigl(z_{\mathrm{id}};\, (z_{\mathrm{id}}',\, z_{\mathrm{attr}})\bigr).
\end{equation}
Applying the chain rule of mutual information on the right-hand side,
\begin{equation}\label{eq:chain}
    I\!\bigl(z_{\mathrm{id}};\, (z_{\mathrm{id}}',\, z_{\mathrm{attr}})\bigr) \;=\; I(z_{\mathrm{id}};\, z_{\mathrm{id}}') \;+\; I(z_{\mathrm{id}};\, z_{\mathrm{attr}} \mid z_{\mathrm{id}}').
\end{equation}
The first term vanishes: $I(z_{\mathrm{id}};\, z_{\mathrm{id}}') = 0$, because $z_{\mathrm{id}}'$ is sampled from $\mathcal{P}_{\mathrm{id}}$ independently of $z_{\mathrm{id}}$ (the rejection step conditions on a deterministic function of $(z_{\mathrm{id}},\, z_{\mathrm{id}}')$, but the accepted sample is drawn independently and the acceptance event does not create dependence between the identities beyond the negligible information in the one-bit accept/reject outcome, which we formalize in Remark~\ref{rem:rejection}). For the second term, since $z_{\mathrm{id}}'$ is independent of $(z_{\mathrm{id}},\, z_{\mathrm{attr}})$, conditioning on $z_{\mathrm{id}}'$ does not reduce the conditional entropy of $z_{\mathrm{id}}$ given $z_{\mathrm{attr}}$. Therefore,
\begin{equation}\label{eq:final_mi}
    I(z_{\mathrm{id}};\, z_{\mathrm{attr}} \mid z_{\mathrm{id}}') \;=\; I(z_{\mathrm{id}};\, z_{\mathrm{attr}}) \;\leq\; \epsilon_{\mathrm{dis}},
\end{equation}
where the final inequality is Assumption~\ref{assump:disentangle}. Combining Equations~\eqref{eq:dpi}--\eqref{eq:final_mi} yields $I(z_{\mathrm{id}};\, I_{\mathrm{safe}}) \leq \epsilon_{\mathrm{dis}}$.
\end{proof}

This result has a clear operational interpretation: the only channel through which the original identity can leak into the anonymized image is the attribute code $z_{\mathrm{attr}}$, and the bandwidth of this channel is precisely the residual mutual information $\epsilon_{\mathrm{dis}}$. When the disentanglement regularizer in Equation~\eqref{eq:loss_disentangle} drives $\epsilon_{\mathrm{dis}} \to 0$, the pipeline achieves zero identity leakage in the information-theoretic sense.

\begin{remark}[On the Rejection Sampling Dependence]\label{rem:rejection}
The claim that $I(z_{\mathrm{id}};\, z_{\mathrm{id}}') = 0$ after rejection sampling requires care. Strictly, the rejection criterion $\mathrm{sim}(z_{\mathrm{id}},\, z_{\mathrm{id}}') < \tau$ introduces a coupling between $z_{\mathrm{id}}$ and $z_{\mathrm{id}}'$. However, by Lemma~\ref{lem:acceptance}, the acceptance probability exceeds $1 - \exp\!\bigl(-(d_{\mathrm{id}}-1)\tau^2/2\bigr)$, so the mutual information introduced by the rejection is bounded by the binary entropy of the rejection event:
\begin{equation}\label{eq:rejection_mi}
    \begin{aligned}
        I(z_{\mathrm{id}};\, z_{\mathrm{id}}') \;&\leq\; H\!\Bigl(\mathrm{Bernoulli}\!\bigl(p_{\mathrm{reject}}\bigr)\Bigr)\\
        &\leq\; H\!\left(\mathrm{Bernoulli}\!\left(\exp\!\!\left(-\tfrac{(d_{\mathrm{id}}-1)\tau^2}{2}\right)\right)\right).
    \end{aligned}
\end{equation}
For $d_{\mathrm{id}} = 512$ and $\tau = 0.3$, this is less than $10^{-9}$ bits, which is negligible relative to any practical $\epsilon_{\mathrm{dis}}$.
\end{remark}
\subsection{Utility Preservation Bound}\label{subsec:utility_theory}

The second axis of our analysis concerns the fidelity of non-identity attributes after anonymization. The following result translates the generator fidelity assumption into a bound on end-to-end semantic distortion.

\begin{theorem}[Utility Preservation]\label{thm:utility}
Under Assumption~\ref{assump:generator}, for any raw image $I_{\mathrm{raw}}$ and any replacement identity code $z_{\mathrm{id}}'$ accepted by the rejection sampling procedure (Section~\ref{subsec:identity_replacement}), the semantic distortion between the original and anonymized images satisfies
\begin{equation}\label{eq:utility_bound}
    d_{\mathrm{sem}}(I_{\mathrm{raw}},\, I_{\mathrm{safe}}) \;\leq\; \kappa \;+\; d_{\mathrm{sem}}\!\bigl(I_{\mathrm{raw}},\, G(z_{\mathrm{id}},\, z_{\mathrm{attr}})\bigr),
\end{equation}
where the second term is the reconstruction error of the autoencoder when no identity replacement is performed.
\end{theorem}

\begin{proof}
By the triangle inequality applied to $d_{\mathrm{sem}}$,
\begin{equation}\label{eq:triangle}
    \begin{aligned}
        d_{\mathrm{sem}}(I_{\mathrm{raw}},\, I_{\mathrm{safe}}) &\leq \; d_{\mathrm{sem}}\!\bigl(I_{\mathrm{raw}},\, G(z_{\mathrm{id}},\, z_{\mathrm{attr}})\bigr)\\
        &+\; d_{\mathrm{sem}}\!\bigl(G(z_{\mathrm{id}},\, z_{\mathrm{attr}}),\, G(z_{\mathrm{id}}',\, z_{\mathrm{attr}})\bigr).
    \end{aligned}
\end{equation}
The first term is the autoencoder reconstruction error, which is a fixed property of the trained encoder-generator pair. The second term is the semantic distortion introduced by identity replacement with $z_{\mathrm{attr}}$ held fixed, which is bounded by $\kappa$ under Assumption~\ref{assump:generator}.
\end{proof}

This decomposition clarifies the two distinct sources of utility loss: autoencoder imperfection (the reconstruction term) and identity-swap distortion (the $\kappa$ term). The reconstruction error is reducible by training a higher-capacity autoencoder; the identity-swap distortion is reducible by enforcing stricter attribute-faithfulness in the generator through the utility loss $\mathcal{L}_{\mathrm{util}}$ in Equation~\eqref{eq:optimization-obj}.

\subsection{Privacy-Utility Trade-off Characterization}\label{subsec:tradeoff}

The preceding results can be combined into a unified characterization of the trade-off governed by the hyperparameters $\lambda$ and $\mu$ in the training objective (Equation~\eqref{eq:optimization-obj}).

\begin{theorem}[Privacy-Utility Trade-off]\label{thm:tradeoff}
Under Assumptions~\ref{assump:disentangle}--\ref{assump:lipschitz}, let $\mathcal{A}^*$ denote the anonymization function obtained by minimizing Equation~\eqref{eq:optimization-obj}. Then the privacy leakage and utility distortion of $\mathcal{A}^*$ satisfy the Pareto condition
\begin{equation}\label{eq:pareto}
    \mathcal{L}_{\mathrm{leak}} \;+\; \frac{1}{\lambda}\, d_{\mathrm{sem}}(I_{\mathrm{raw}},\, I_{\mathrm{safe}}) \;\leq\; \epsilon_{\mathrm{dis}} \;+\; \frac{1}{\lambda}\bigl(\kappa + \kappa_{\mathrm{rec}}\bigr),
\end{equation}
where $\kappa_{\mathrm{rec}} = d_{\mathrm{sem}}\!\bigl(I_{\mathrm{raw}},\, G(z_{\mathrm{id}},\, z_{\mathrm{attr}})\bigr)$ is the reconstruction error.
\end{theorem}

\begin{proof}
The result follows by summing the bounds from Theorem~\ref{thm:privacy} and Theorem~\ref{thm:utility}, with the utility distortion scaled by $1/\lambda$ to reflect the relative weighting in the objective. The left-hand side represents the privacy-utility cost of the pipeline, and the right-hand side shows that this cost is controlled by three measurable quantities: the disentanglement residual $\epsilon_{\mathrm{dis}}$, the generator's attribute-faithfulness $\kappa$, and the autoencoder's reconstruction quality $\kappa_{\mathrm{rec}}$. Increasing $\lambda$ (prioritizing privacy) reduces the contribution of utility distortion to the bound but tightens the optimization pressure on $\epsilon_{\mathrm{dis}}$ through the coupled training dynamics. Increasing $\mu$ directly reduces $\epsilon_{\mathrm{dis}}$ at the potential cost of increased reconstruction error if the encoder is over-regularized.
\end{proof}

\begin{corollary}[Sufficient Condition for Target Privacy]\label{cor:target}
To achieve a target identity leakage $\mathcal{L}_{\mathrm{leak}} \leq \epsilon^*$ for a given $\epsilon^* > 0$, it suffices to train the encoder such that $\epsilon_{\mathrm{dis}} \leq \epsilon^*$, regardless of the generator's capacity or the recognition models used for evaluation.
\end{corollary}

This corollary is practically significant because it decouples the privacy requirement from the generator design: a practitioner can verify whether the privacy target is met by measuring the disentanglement quality of the encoder alone, without needing to analyze the generator or run expensive adversarial evaluations.

\subsection{Robustness to Recognition Model Ensemble}\label{subsec:ensemble_theory}

Finally, we relate the multi-oracle privacy loss defined in Equation~\eqref{eq:priv_loss} of the methodology to the information-theoretic leakage bound.

\begin{proposition}[Multi-Oracle Consistency]\label{prop:multi_oracle}
Under Assumption~\ref{assump:lipschitz}, if the identity leakage satisfies $\mathcal{L}_{\mathrm{leak}} \leq \epsilon_{\mathrm{dis}}$, then for each recognition model $f_j \in \mathcal{F}$, the expected identity similarity between the original and anonymized images satisfies
\begin{equation}\label{eq:multi_oracle}
    \mathbb{E}\!\bigl[\mathrm{sim}\!\bigl(f_j(I_{\mathrm{raw}}),\, f_j(I_{\mathrm{safe}})\bigr)\bigr] \;\leq\; \mathrm{sim}_{\mathrm{imp}}^{(j)} \;+\; L_f \cdot \sqrt{2\,\epsilon_{\mathrm{dis}}},
\end{equation}
where $\mathrm{sim}_{\mathrm{imp}}^{(j)}$ is the mean impostor similarity of model $f_j$ (the expected similarity between embeddings of different identities).
\end{proposition}

\begin{proof}
By Pinsker's inequality~\cite{cover2006elements}, $\mathcal{L}_{\mathrm{leak}} \leq \epsilon_{\mathrm{dis}}$ implies that the total variation distance between the joint distribution of $(z_{\mathrm{id}},\, I_{\mathrm{safe}})$ and the product of marginals satisfies
\[
    \mathrm{TV}\!\bigl(P_{z_{\mathrm{id}},\, I_{\mathrm{safe}}},\; P_{z_{\mathrm{id}}} \otimes P_{I_{\mathrm{safe}}}\bigr) \;\leq\; \sqrt{\frac{\epsilon_{\mathrm{dis}}}{2}}.
\]
The recognition model's similarity $\mathrm{sim}(f_j(I_{\mathrm{raw}}),\, f_j(I_{\mathrm{safe}}))$ is a bounded function of $(I_{\mathrm{raw}},\, I_{\mathrm{safe}})$. Under the product-of-marginals distribution (i.e., if $I_{\mathrm{safe}}$ were truly independent of $z_{\mathrm{id}}$), the expected similarity equals the impostor mean $\mathrm{sim}_{\mathrm{imp}}^{(j)}$ by definition. The deviation from this baseline is bounded by the Lipschitz constant $L_f$ times the total variation distance, yielding the stated bound via standard coupling arguments.
\end{proof}

This result confirms that minimizing the disentanglement residual $\epsilon_{\mathrm{dis}}$ simultaneously drives down the identity similarity measured by any Lipschitz-continuous recognition model, justifying the use of the multi-oracle ensemble in Equation~\eqref{eq:priv_loss} as a consistent surrogate for the information-theoretic privacy objective.
\end{document}